\documentclass[11pt,a4paper]{article}
\usepackage[hyperref]{emnlp2020}
\usepackage{times}
\usepackage{latexsym}
\usepackage{pslatex}
\usepackage[english]{babel}
\usepackage[utf8]{inputenc}
\usepackage{amsmath}
\usepackage{bm}
\usepackage{graphicx}
\usepackage{xcolor}
\usepackage{tikz}
\usepackage{url}
\usepackage{float}
\usepackage{rotating}
\usepackage{natbib}
\usepackage{amssymb}
\usepackage{linguex}
\usepackage{xspace}
\usepackage{pgf}
\usepackage{colortbl}
\usepackage{multirow}
\usepackage{array}
\usepackage{booktabs}

\newcolumntype{M}[1]{>{\centering\arraybackslash}m{#1}}

\newcommand\blfootnote[1]{%
  \begingroup
  \renewcommand\thefootnote{}\footnote{#1}%
  \addtocounter{footnote}{-1}%
  \endgroup
}

\usepackage{microtype}

\aclfinalcopy 

\title{Structural Supervision Improves Few-Shot Learning and Syntactic Generalization in Neural Language Models}

\author{Ethan Wilcox$^1$, Peng Qian$^2$, Richard Futrell$^3$, Ryosuke Kohita$^4$, Roger Levy$^2$ and Miguel Ballesteros$^5$ \\

  $^1$Department of Linguistics, Harvard University  \\
  $^2$Department of Brain and Cognitive Sciences, Massachusetts Institute of Technology\\
  $^3$Department of Language Science, UC Irvine
  $^4$IBM Research 
  $^5$Amazon AI \\
  \texttt{wilcoxeg@g.harvard.edu}, \texttt{\{pqian,rplevy\}@mit.edu}\\
  \texttt{rfutrell@uci.edu}, \texttt{kohi@ibm.com}, \texttt{ballemig@amazon.com} } 

\date{}

\begin{document}

\setlength{\Exlabelwidth}{0.7em}
\setlength{\Exlabelsep}{0.7em}
\setlength{\SubExleftmargin}{1.3em}
\setlength{\Extopsep}{2pt}

\maketitle
\begin{abstract}
Humans can learn structural properties about a word from minimal experience, and deploy their learned syntactic representations uniformly in different grammatical contexts. We assess the ability of modern neural language models to reproduce this behavior in English and evaluate the effect of structural supervision on learning outcomes. First, we assess few-shot learning capabilities by developing controlled experiments that probe models' syntactic nominal number and verbal argument structure generalizations for tokens seen as few as two times during training. Second, we assess invariance properties of learned representation: the ability of a model to transfer syntactic generalizations from a base context ({\it e.g.}, a simple declarative active-voice sentence) to a transformed context ({\it e.g.}, an interrogative sentence). We test four models trained on the same dataset: an $n$-gram baseline, an LSTM, and two LSTM-variants trained with explicit structural supervision \citep{dyer2016rnng, charniak2016parsing}. We find that in most cases, the neural models are able to induce the proper syntactic generalizations after minimal exposure, often from just two examples during training, and that the two structurally supervised models generalize more accurately than the LSTM model. All neural models are able to leverage information learned in base contexts to drive expectations in transformed contexts, indicating that they have learned some invariance properties of syntax.\blfootnote{Miguel conducted this work while at IBM Research} 
\end{abstract}
 
\section{Introduction}

\renewcommand{\thefootnote}{}
\footnotetext{
Scripts and data for this paper can be found online at \texttt{https://github.com/wilcoxeg/fsl\_invar}}
\renewcommand{\thefootnote}{\arabic{footnote}}

Recurrent Neural Network language models \cite{elman1990finding, hochreiter1997long} have been shown to learn many aspects of natural language syntax including a number of long-distance dependencies and representations of incremental syntactic state \cite{marvin2018targeted, wilcox2018rnn, futrell2018rnns}. However, previous studies have not investigated the relationship between a token's frequency in the training corpus and syntactic properties models learn about it. In this work, we assess neural models' ability to make robust syntactic generalizations about a token's nominal number or verbal argument structure based on minimal exposure with the token during training. Because of the Zipfian distribution of words in a corpus, the vast majority of word types will be seen only a handful of times during training \cite{zipf1949human}. Therefore, the few-shot learning capabilities of neural LMs are critical to their robustness as an NLP system and as a cognitive model.

However, human learning goes beyond simply learning syntactic properties in particular constructions. People apply the same properties across different constructions, meaning that their representations of the syntactic features of a word are in some sense \emph{invariant} to the grammatical context of that word. For example, speakers and listeners are sensitive to a verb's argument structure relationships and can easily recognize that a verb which cannot take a direct object in active, declarative sentences cannot be passivized (as in the ungrammatical sentence \textit{``The ship was arrived."}) The relationship between an active sentence and a passive sentence has been termed a \textit{transformation} in the linguistic literature \citep{chomsky1957syntactic}. Many semantic-syntactic rules that govern word co-occurrence in one form, such as a verb's argument structure relationships, hold uniformly across transformations. It remains an open question whether models learn grammatical rules invariant to their surface realization, a property we call \textit{syntactic invariance}.

We combine assessment of few-shot learning and syntactic invariance for two grammatical features of English: whether a noun is singular or plural (nominal number) and whether a verb is transitive or intransitive (verbal argument structure). We assess whether a model is able to make different predictions based on number or argument structure in a simple active voice \textit{base context}. We then assess whether models are able to make similar distinctions in a \textit{transformed context}---passive voice for verbs and polar questions for nouns. In the \textit{transformed contexts}, we test models with tokens that occur only in the \textit{base context} during training. (As a control, we also test verbal passive voice with tokens that occur in the passive voice in the training data to establish that models have learned the proper syntactic rules for this context.) For models to succeed in the transformed contexts they must represent syntactic features in a way that is invariant to the specific realization of those features in terms of word co-occurrences in different constructions. For each grammatical feature, we introduce a suite of novel targeted test sentences, similar to those presented in \citet{marvin2018targeted}.

We find that all neural models tested are able to induce the proper syntactic generalizations in the \textit{base} and \textit{transformed} contexts after just two or three exposures, whereas a baseline $n$-gram model fails to learn the relevant generalizations. For all constructions tested our two neural models enhanced with explicit structural supervision outperform the purely sequence model. Assessing invariance properties, we find that neural models demonstrate proper behavior in transformed contexts, even for tokens seen only in base contexts during training. This behavior indicates that models are able to deploy generalizations learned in one syntactic context into different syntactic environments, a key component of human linguistic capabilities that has been so far untested in the neural setting.

\subsection{Related Work}

Bayesian models of word learning have shown successes in acquiring proper syntactic generalizations from minimal exposure \citep{tenenbaum2000word,wang-etal:2017-distributional}, however it is not clear how well neural network models would exhibit these rapid generalizations. Comparing between neural network architectures, recent work has shown that models enhanced with explicit structural supervision during training produce more humanlike syntactic generalizations \cite{kuncoro2016recurrent, kuncoro2018lstms, wilcox2019structural}, but it remains untested whether such supervision helps learn properties of tokens that occur rarely during training.

Previous studies have found that Artificial Neural Networks (ANNs) are capable of learning some argument structure paradigms and make correct predictions across multiple frames \cite{kann2018verb}, however these capabilities remain untested for incremental language models. Much has been written about the ability of ANNs to learn number agreement \cite{linzen2016assessing, gulordava2018colorless, giulianelli2018under}, including their ability to maintain the dependency across different types of intervening material \cite{marvin2018targeted} and with coordinated noun phrases \cite{an2019representation}. \citet{hu2020systematic} find that model architecture, rather than training data size, may contribute most to performance on number agreement and related tasks. Focusing on RNN models, \citet{lakretz2019emergence} find evidence that number agreement is tracked by specific ``number" units that work in concert with units that carry more general syntactic information like tree depth. \citet{jumelet2019analysing} argue that when learning dependencies RNNs acquire a default form (which they postulate to be singular and masculine), and predicting a non-default form requires explicit contrary evidence. Our results support their hypothesis. Models are more accurate with singular nouns and transitive verbs seen only a few times in training, behavior that indicates these forms are expected when evidence is sparse.

\section{General Methods}

\begin{figure*}[]
\centering
\begin{minipage}{0.98\textwidth}
\centering
\includegraphics[width=\textwidth]{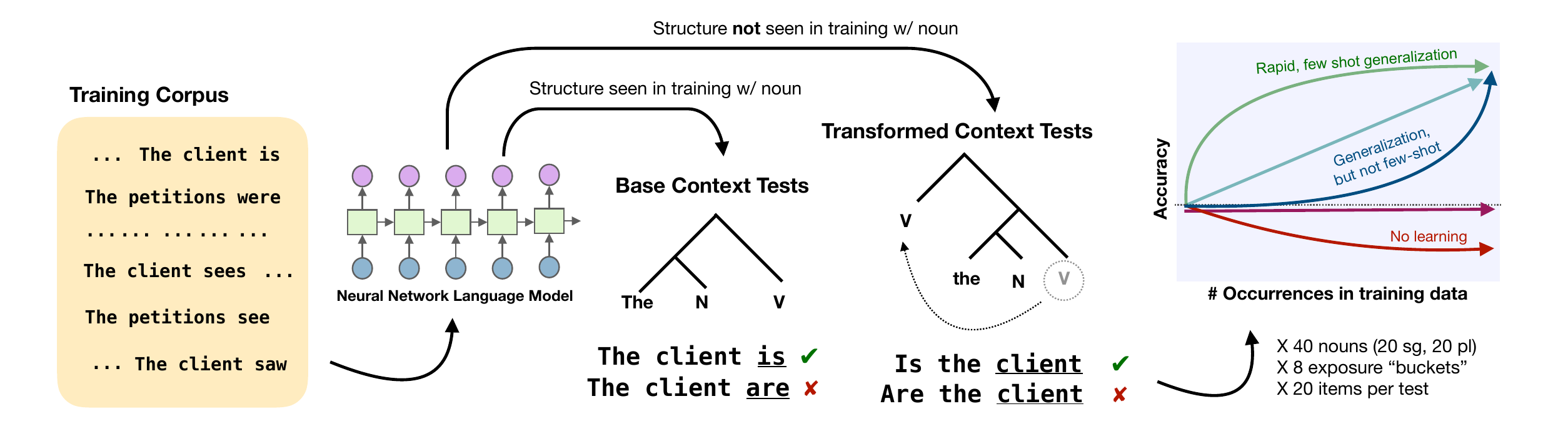}
\end{minipage}
\caption{Experimental pipeline, given for Nominal Number tests without any modification.}
\label{fig:pipeline}
\end{figure*}

\subsection{Psycholinguistic Assessment Paradigm} \label{sec:psycho-methods}

In order to assess the learning outcomes of neural LMs, we adopt the Psycholinguistic Assessment Paradigm \cite{linzen2016assessing, futrell2018rnns}. In this paradigm models are exposed to sentences that reveal the syntactic generalizations learned during training. For example, \citet{linzen2016assessing} used as input sentences with the prefix \textit{The keys to the cabinet \dots} and inspected the relative probabilities of the continuations \textit{is} and \textit{are}. If the model has learned the grammatical rule that the head of a subject noun phrase governs plural agreement, then P(\textit{are}) should be greater than P(\textit{is}). 

In order to assess the few-shot learning capabilities of the models tested, we sample words from eight ``exposure buckets" based on the number of times the word appears in the training corpus, with the majority of buckets for words seen less than 10 times during training.\footnote{Our exposure buckets were: 2, 3, 4, 5, 6-10, 11-20, 21-30, 50-100, and we use the end of the range to label the buckets in figures. Thus, if a token is in exposure bucket ``4" it occurred four times in the training corpus; if it is in exposure bucket ``20," it occurred between 11 and 20 times in the training corpus.} For each test, we sample 40 words balanced across syntactic categories. For each word, we generate 20 test sentences---each with a grammatical and ungrammatical condition---and measure the proportion of sentences in which the model shows higher probability in the grammatical condition in a particular critical region. In total, each test contains 12,800 sentences.

For each experiment, we report two metrics: First, we report the number of exposure buckets for which the models' accuracy is significantly above chance, which is 50\% in all cases.
Second, in order to assess the impact of structural supervision we report the results of a logistic regression model trained to predict accuracy with fixed effects of model class and exposure and random by-sentence intercepts.\footnote{The R code used to run the model was \texttt{glm(accuracy $\sim$ model + exposure-bucket + (1|sentence-id))}} A significant positive effect of model class means that the architecture contributes to more robust learning across all exposure buckets tested.

\subsection{Models Tested}

All models are trained on sections 2-21 of the Wall Street Journal portion of the Penn Treebank (PTB) \cite{marcus1993building}, which contains $\sim$1,000,000 tokens of newswire text. While this dataset is relatively small compared to the ones on which neural models were trained in \citet{linzen2016assessing} and \citet{gulordava2018colorless}, it was imperative that we collect accurate metadata for all tokens in the training data and PTB is one of the largest datasets expertly hand annotated with both syntactic structure and part-of-speech information. Argument structure statistics were obtained from a Universal Dependency representation of the dataset, converted from its original phrase structure parse via the Stanford Parser \cite{schuster2016enhanced}. Additional argument structure information was collected from the Celex2 dataset \citep{baayen1995celex}.

\noindent \textbf{$n$-Gram Baseline} We used a 5-gram baseline with modified Kneser-Ney smoothing trained using SRILM \cite{stolcke2002srilm}.

\noindent {\bf Recurrent Neural Network LMs} model a sentence in a purely sequential basis, without explicitly representing the latent syntactic structure. We use the LSTM architecture \cite{hochreiter1997long} and, following \citet{futrell2018rnns} derive the word surprisal from the LSTM language model by directly computing the negative log value of the predicted conditional probability from the softmax layer. This and subsequent neural models were trained with embedding size 256, dropout 0.3 following the hyper-parameters in \citet{van2018neural}.

\noindent {\bf Recurrent Neural Network Grammars (RNNGs)} \cite{dyer2016rnng} jointly model a sentence as well as its syntactic parse. The model explicitly represents parse trees and composes partially built phrase structures, an approach that may result in better performance on tree-structurally local but linearly distal relationships (see \cite{dyer2016rnng}). Models are supervised with Penn-Treebank style parses during training; we assess whether this explicit syntactic supervision translates into better few-shot learning and syntactic invariance outcomes. We use the same hyperparameters used by \citet{dyer2016rnng}.

\noindent \textbf{ActionLSTM Model:} We ablate the composition function of an RNNG, producing a model that predicts the \textbf{action sequence} of a parse tree as well as the upcoming word. In this sense, it is an incrementalized version of the Parsing-as-Language-Modeling configuration presented in \citet{charniak2016parsing}. We refer to this model as the ``ActionLSTM" model in the following sections.\footnote{As RNNG and ActionLSTM jointly model terminal words and syntactic parses, we use word-synchronous beam search \cite{stern2017effective} to compute surprisal values incrementally. ActionLSTM was able to achieve a parsing F1 score of 92.81 on the PTB, which is in the same range as the original architecture on the same test set, as reported in \citet{kuncoro2016recurrent}.}

\section{Nominal Number}

In English, the matrix verb of a tensed clause must agree in number with the head of the subject Noun Phrase. Neural LMs are capable of learning this relationship based on a pure language modeling objective in multiple languages and across a variety of intervening material \cite{gulordava2018colorless, marvin2018targeted}. Previous work assessing neural learning of subject-verb number agreement has not directly compared the learning outcome to the amount of experience models receive during training.

\begin{figure*}[]
\centering
\begin{minipage}{0.48\textwidth}
\centering
\includegraphics[width=\textwidth]{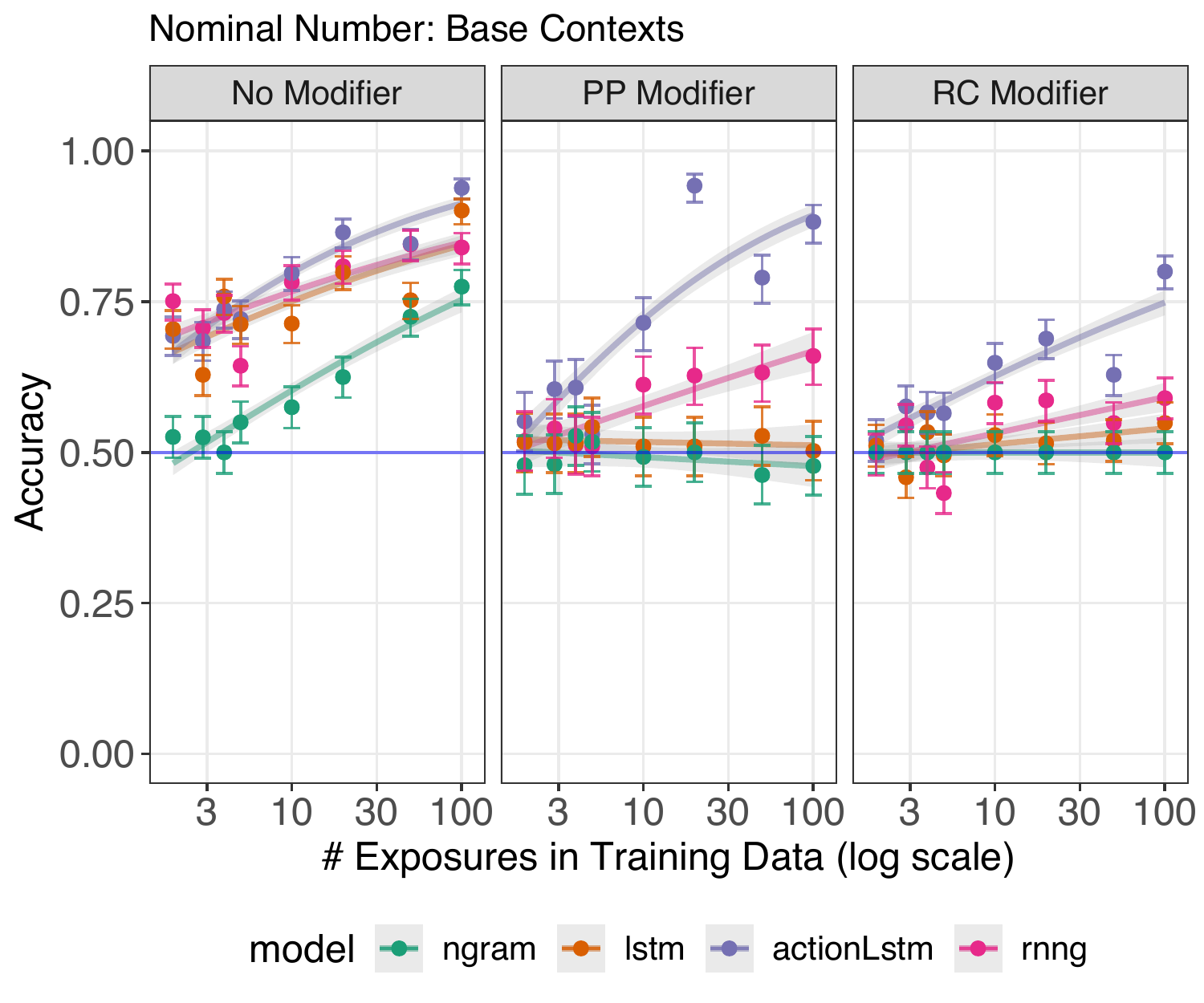}
\end{minipage}
\begin{minipage}{0.48\textwidth}
\centering
\includegraphics[width=\textwidth]{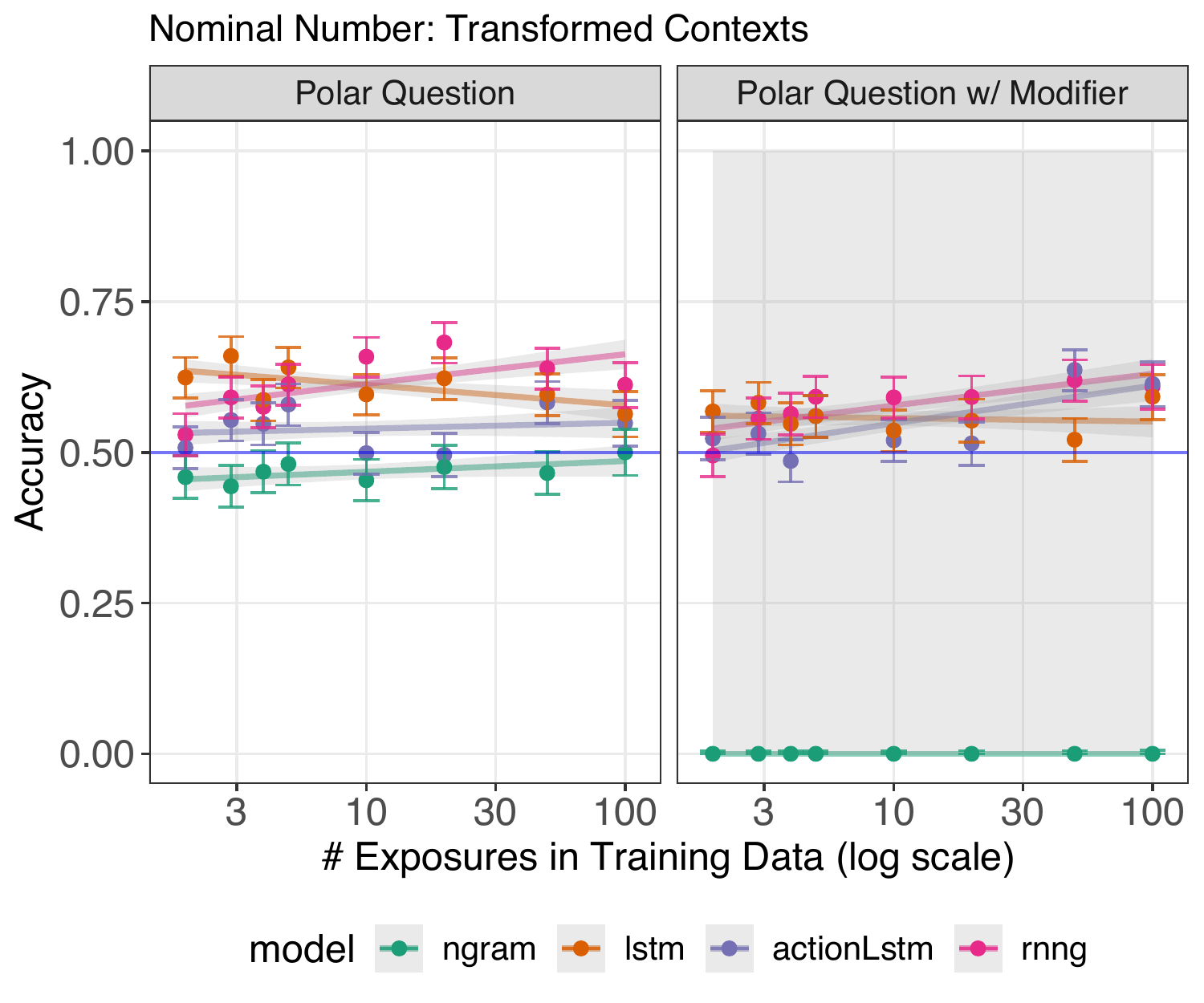}
\end{minipage}
\caption{Results from the novel word learning experiments for Base Contexts (left) and Transformed Contexts (right). Results are averaged across singular and plural nouns. Points represent by-exposure bucket means, with 95\% binomial confidence intervals. Smooth lines are results of logistic regression model fits on the raw data.}
\label{fig:number-results}
\end{figure*}

\subsection{Base Contexts: Active Voice}

In order to assess the few-shot learning capabilities of neural models in base contexts, we randomly select 20 plural and 20 singular nouns in each exposure bucket and generated test items for each following \ref{ex:num-base-simple}. (In \ref{ex:num-base-simple} and following examples, un-grammatical sentence variants are marked with a *, which is for presentational purposes only and not included in test items. Underlined portions of the sentences indicate critical regions, whose joint probability was used to calculate accuracy scores.) To test whether models' representations are impervious to modification, we also generate a set of test items with prepositional phrases (PPs) and object relative clauses (ORCs) modifying the head verb, following \ref{ex:num-base-pp} and \ref{ex:num-base-orc}. In this and all following experiments, sentences are generated using templates and---other than the target token---contain tokens that occur at least 50 times in the training data. As described in \ref{sec:psycho-methods}, strength of nominal number feature learning was evaluated by calculating model accuracy, or the proportion of times the models preferred the grammatical variant.

\ex. \textbf{Base No Modifier} (singular example) \label{ex:num-base-simple}
\a. The president \underline{is}... 
\b. *The president \underline{are}...

\ex. \textbf{Base w/ PP Modifier} (plural example) \label{ex:num-base-pp}
\a. *The petitions near the old investment \underline{is}... 
\b. The petitions near the old investment \underline{are}... 

\ex. \textbf{Base w/ ORC Modifier} (singular example) \label{ex:num-base-orc}
\a. The client that the lawyers like \underline{is}...
\b. *The client that the lawyers like \underline{are}...

The results for this experiment can be seen in Figure~\ref{fig:number-results}, in the left panels. This is the same presentational paradigm we will use for all results in this paper: the $y$-axis is the model's accuracy, pooled across performance on singular and plural nouns, and the $x$-axis is the model's exposure for each token---the number of times it occurs during training. The scale of the x-axis is log-transformed. Points represent mean accuracy for each exposure bucket and error bars are 95\% binomial confidence intervals. Lines show logistic regression fits from the raw data, with standard errors.

In the \textit{Base Simple} experiment, the $n$-gram shows moderate few-shot learning, above chance in 5/8 exposure buckets. All neural models show robust few-shot learning and are above chance in all exposure buckets. We find a significant effect of structural supervision, with both the ActionLSTM and RNNG outperforming the LSTM model ($p<0.05$ and $p<0.001$ respectively). The experiments with modifiers prove more difficult: In the \textit{Base PP} experiment, we find no few-shot generalization for the $n$-gram model, weak few-shot generalization for the LSTM (1/8 buckets), moderate generalization for the RNNG (4/8 buckets) and strong generalization for the ActionLSTM (7/8 buckets). In the \textit{Base RC} experiments we find a similar pattern: No generalization for the $n$-gram (0/8 buckets), weak generalization for the LSTM (2/8 buckets), but stronger generalization for the two structurally supervised models (6/8 and 7/8 buckets for the RNNG and ActionLSTM respectively). For these two experiments, we find an effect of structural supervision on accuracy, with both the ActionLSTM and RNNG out-performing the LSTM ($p<0.001$ except for the RNNG RC-Modifier where $p<0.05$). Our results are generally in line with those presented in \citet{marvin2018targeted}, who find performance in the 50-60\% accuracy range for number agreement across PP and RC modifiers. Some studies, such as \citet{lakretz2019emergence} find higher performance performance on a similar task; we attribute these differences to the relative size of the training data. 

Overall, these results indicate that all models are capable of making grammatical generalizations based on minimal exposure with a token, and capable of leveraging increased exposure to make more accurate number agreement predictions. (For statistical analysis of the effects of exposure on accuracy across all of our experiments, see Appendix A.) Although the graphs in Figure \ref{fig:number-results} are pooled across noun category, models demonstrate much higher accuracy for \textit{singular} nouns, especially if they occur only a few times in training. Improvement from increased exposure comes as models get better at accurately predicting number agreement for plural nouns. These findings are in line with the singular bias reported in \citet{marvin2018targeted} and support the hypothesis from \citet{jumelet2019analysing} that models have learned a ``default" prediction, in this case \textit{singular}. For results and analysis from all of our experiments broken down by grammatical category, see Appendix B.

\subsection{Transformed Contexts \& Syntactic Invariance}

Because subjects precede verbs in English, most evidence available to a neural model for a noun's number follows the noun linearly. However, in English \textit{polar question} formation, the matrix verb is moved to the front of the sentence inverting the base noun-verb order, as in \ref{ex:number-transf-simple}. If models have learned nominal number feature that is invariant to linear order, verbs that set off polar questions should set up expectations for nominal subjects that match in number. In order to assess whether models were robust to such transformations, we created test items following the template in \ref{ex:number-transf-simple} and \ref{ex:number-transf-mod}, which includes an additional four word modification. Half of the sentences were in present tense, half in past tense. We measure the model's accuracy at the noun directly and pool accuracy scores across singular and plural nouns.

\ex. \textbf{Polar Question} (singular example) \label{ex:number-transf-simple}
\a. Is the \underline{president}... 
\b. * Are the \underline{president}... 

\ex. \textbf{Polar Question w/ Modifier} (plural example) \label{ex:number-transf-mod}
\a. * Is the very big and important \underline{hearings}... 
\b. Are the very big and important \underline{hearings}...

Although polar questions are relativally rare in the WSJ section of the Penn Treebank---the ratio of active to inverted polar sentences is $\sim$1000:1---some nouns do occur in both base for and inverted form. Because our aim here is to assess models' generalization to novel syntactic frames, we filtered every noun from our previous set that occurred in both frames, a total of 15 nouns. The results presented here therefore address whether the models have learned a representation of number that is invariant to linear order. Successful learning, in this case, means that models have learned that nouns which set up expectations for singular verbal inflections should also be more likely in contexts where singular nouns are expected, and likewise for plurals.

The results for this experiment can be seen in Figure \ref{fig:number-results} in the right-hand panel. In the \textit{Transformed Simple} experiment we find no generalization for the $n$-gram model, which was not above chance in any exposure bucket, moderate generalization for the ActionLSTM (above chance in 5/8 buckets) and strong generalization for the LSTM and RNNG model (8/8 buckets and 7/8 buckets respectively). Results are similar for the \textit{Transformed Modifier} experiment. In this case the $n$-gram model is at 0\% accuracy for all buckets; this is because it assigns equal probability to the critical region in each condition, which we count as a ``failure." The ActionLSTM displays moderate generalization (4/8 buckets) and the RNNG and LSTM stronger generalization (7/8 and 8/8 buckets respectively). In these experiments, the ActionLSTM under-performs compared to the LSTM in the \textit{No Modifier} experiment ($p<0.001$), but the RNNG outperforms it in the \textit{Transformed Modifier} experiment ($p<0.05$). These results indicate that all the neural models are able to leverage information gained in the base contexts to drive expectations in the inverted context, however note that the accuracy scores are lower here than in the base contexts, with no model breaking the 70\% accuracy threshold. This may be due to the relatively few number of polar questions in the corpus.

\section{Verbal Argument Structure}

In this section, we assess the ability of neural models to represent verbal argument structure, which we simplify to whether a verb is transitive or intransitive. If it is transitive, then it requires a theme, which must be realized as a direct object in the active voice. If it is intransitive then it requires an empty theme position and cannot have an object in the active voice. Verbal argument structure is a hard task, insofar as both intransitivity and transitivity can only be inferred through \textit{indirect negative evidence}. We assess neural models' sensitivity for indirect negative evidence by investigating how much experience models need with a particular verb before they make robust predictions about whether an object should follow that token at test time.

\begin{figure*}[]
\centering

\begin{minipage}{0.48\textwidth}
\centering
\includegraphics[width=\textwidth]{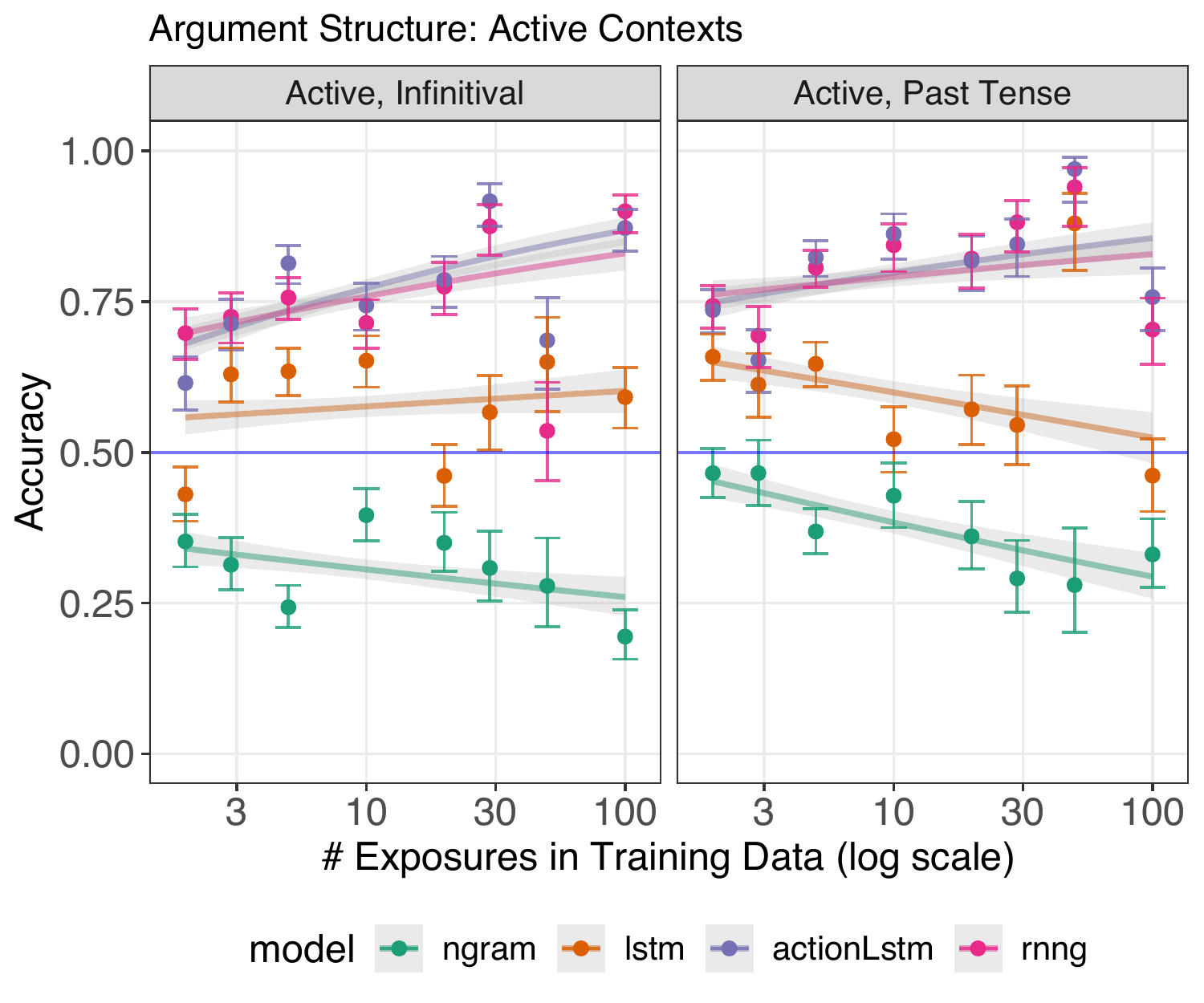}
\end{minipage}
\begin{minipage}{0.48\textwidth}
\centering
\includegraphics[width=\textwidth]{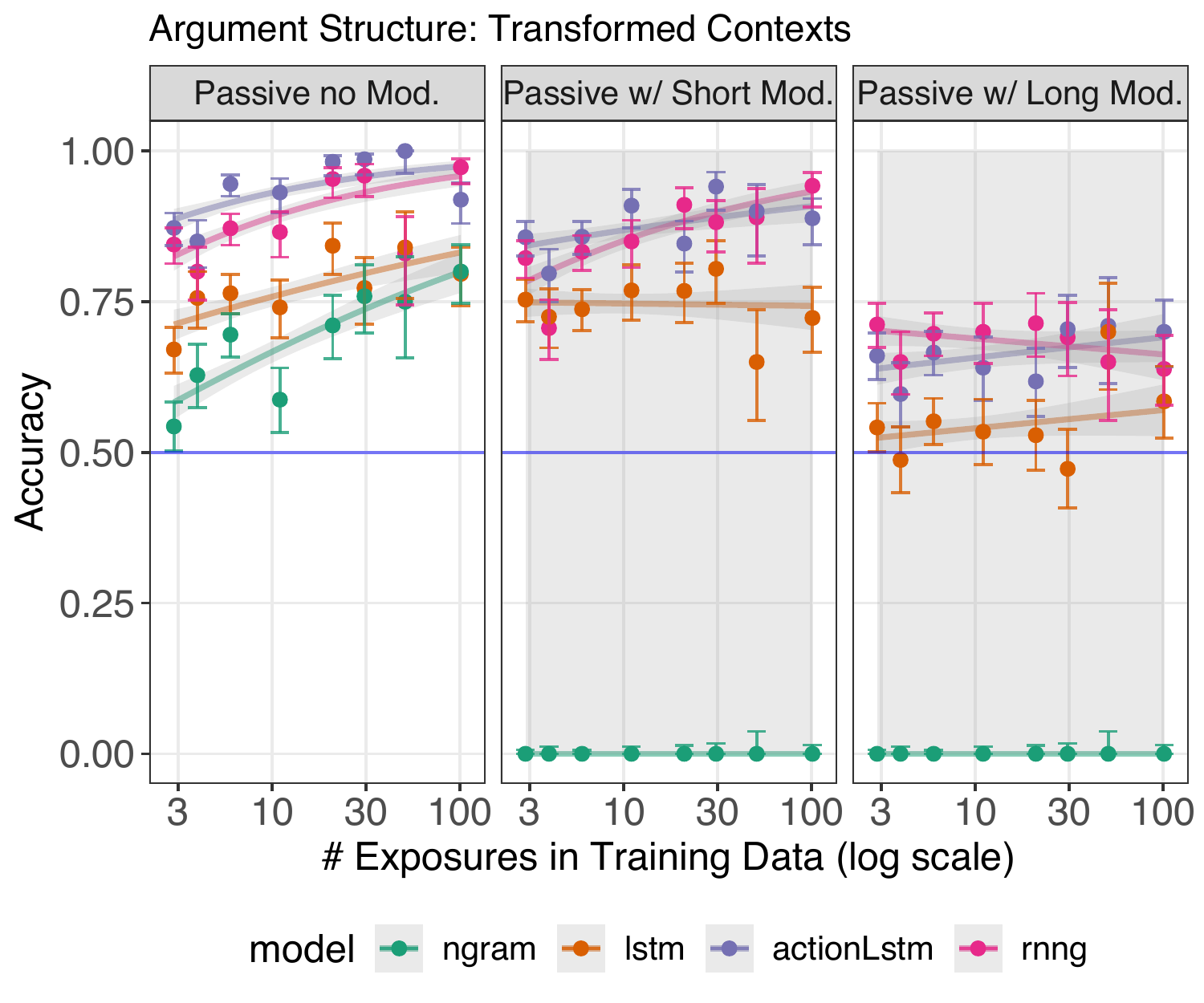}
\end{minipage}
\caption{Results from the few shot learning experiments for active contexts (left) and transformed, passive, contexts (right). Points represent mean accuracy for each exposure and error bars are 95\% binomial confidence intervals. Lines show logistic regression fits from the raw data, with standard errors.}
\label{fig:argstruct-results}
\end{figure*}

\subsection{Base Contexts: Active Voice} \label{sec:argstruct-base}

In order to assess the generalizations models have learned about verbal argument structure, we selected 20 transitive verbs and 20 intransitive verbs. Verb transitivity was assessed using the hand-coded Celex2 Corpus, and double-checked using a Universal Dependencies representation of the original PTB Phrase Structure trees; verbs marked ``transitive'' in Celex were dropped if they occurred without arguments in 3rd person past tense active voice less than 90\% of the time and verbs marked ``intransitive'' in Celex were dropped if they took arguments in 3rd person past tense more than 10\% of the time.

We generated sentences for each verb in an \textit{infinitival} construction, following \ref{ex:argstruct-base-pres} (which gives a transitive example) and in \textit{past tense} following \ref{ex:argstruct-base-past} (which gives an intransitive example). Infinitival tense sentences were generated because most verbs are ambiguous between their past tense and past participle forms, which can occur without a direct object, even for transitive verbs (e.g. \textit{The patient was cured}). Model accuracy was assessed by comparing the surprisal of the \texttt{adverb + period} region at the end of each sentence (e.g. ``today ." in the examples below) and accuracy scores are pooled across verb type. For transitive verbs, this region should be more surprising if an object is absent; for intransitive verbs this region should be more surprising if it is present.

\ex. \textbf{Active, Infinitival Tense} (transitive example) \label{ex:argstruct-base-pres}
\a. The doctor can cure the patient \underline{today.}
\b. * The doctor can cure \underline{today.}

\ex. \textbf{Active, Past Tense} (intransitive example) \label{ex:argstruct-base-past}
\a. * The doctor slept the patient \underline{today.}
\b. The doctor slept \underline{today.}

The results for this experiment can be seen in Figure \ref{fig:argstruct-results} on the left, with the \textit{infinitival} experiment on the far left and the \textit{past tense} experiment in the middle. For the \textit{Active Infinitival} results we find no generalization for the $n$-gram model, but strong generalization for the three neural models: The LSTM is significantly above chance in 6/8 exposure buckets, the ActionLSTM in 8/8 and the RNNG in 7/8. For the \textit{Active Past} experiment we find no few-shot learning for the $n$-gram model (it is above chance in 0/8 buckets), moderate few-shot learning for the LSTM model (5/8 buckets) but strong few-shot learning for the two supervised models (8/8 buckets for both). We find a significant effect of structural supervision, whereby the ActionLSTM and RNNG outperform the LSTM in both experiments ($p<0.001$). These results indicate that not only that all neural models have learned the basic facts of argument structure, but that they are willing to generalize about the likelihood of an upcoming object after just two exposures during training.

Interestingly, the LSTM model shows a decrease in accuracy in our Past Tense test as words grow more frequent in the training data. We hypothesized that this is because high-frequency tokens ending in ``-ed" are more likely to be used in passive voice, where they occur without a direct object. Models that lack POS disambiguation, would erroneously come to expect these tokens to occur without an object, even when they are being used in active voice, which explains why we do not see the trend for infinitival tests, nor for the supervised models which predict POS tags. We confirmed our hypothesis with two follow-up statistical tests: We found a positive correlation between a token's frequency and the percentage of time it is used in passive voice in the training data ($cor=0.39; p<0.001$). And we ran a statistical model looking at the effect of passive usage on accuracy, finding a positive effect for the two supervised models and a negative effect for the LSTM ($p<0.001$ in all cases).\footnote{The statistical model used was \texttt{glm(accuracy $\sim$ percent\_VBN + (1|sentence\_id)}, where the predictor \texttt{percent\_VBN} indicates the proportion of total occurrences the token is tagged as a passive participle.}

\subsection{Transformed Context: Passive Voice} \label{sec:argstruct-passive}

When verbs are realized in different syntactic frames, or \textit{syntactic transformations}, their argument structure properties are preserved. For example, because passive voice promotes the semantic \textit{theme}---which plays the syntactic role of object in active voice---to the subject position, it is impossible for a truly intransitive verb to be realized in passive voice, as in the ungrammatical \ref{ex:argstruct-passive-ungram}.

In order to assess whether models have learned the rules governing the passive transformation, we designed tests with the same verbs as in Section \ref{sec:argstruct-base}). Verbs were dropped if they had different forms for 3rd person past-tense and passive participle, such as the verb \textit{give} (\textit{gave}, \textit{given}). We generated items following three tests sketched in \ref{ex:argstruct-transf-nomod}, \ref{ex:argstruct-transf-mod} and \ref{ex:argstruct-transf-longmod}. Model accuracy was assessed by comparing the surprisal of the \texttt{verb + adverb + period} following a prefix that contains a passive ``was" versus a prefix that does not, and accuracy scores are pooled across transitive and intransitive verbs. If models are learning the proper grammatical generalizations, then intransitive verbs should be unexpected in passive voice, and the verb should be more surprising when it follows the passivizing ``was". Conversely, transitive verbs should be more likely in passive voice than in active voice without a direct object, which is ruled out by the \texttt{adverb + period} portion of our continuation, and therefore should be more likely when the passivizing ``was" is absent. Because we use verbs that do appear in passive voice during training this section does not test models' invariance properties, but rather their few-shot learning capabilities for this grammatical context. After these control experiments, we turn to invariance tests in Section \ref{sec:argstruct_invar}.

\ex. \textbf{Passive Voice: No Modifier} \label{ex:argstruct-transf-nomod}\\
\textit{Example with Transitive verb}
\a. The doctor was \underline{cured yesterday .}
\b. * The doctor \underline{cured yesterday .}\\
\noindent \textit{Example with Intransitive Verb}
\c.* The doctor was \underline{arrived yesterday .} \label{ex:argstruct-passive-ungram}
\d. The doctor \underline{arrived yesterday.}

\ex. \textbf{Passive Voice: Short Modifier} \label{ex:argstruct-transf-mod}
\a.The dog was quickly and fully \underline{cured today.}
\b. * The dog quickly and fully \underline{cured today.}

\ex. \textbf{  Passive Voice: Long Modifier} \label{ex:argstruct-transf-longmod}
\a. The dog was quickly, suddenly, and entirely \underline{cured yesterday.}
\b. * The dog quickly, suddenly, and entirely \underline{cured yesterday.}

The results for this experiment can be seen in Figure \ref{fig:argstruct-results}, in the right panel. For the \textit{Passive No-Modifier} test, all models are significantly above chance for all exposure buckets tested. For the \textit{Short Modifier} test, we find no few-shot learning for the $n$-gram model (above chance in 0/8 buckets), but strong few-shot learning for all neural models tested (8/8 buckets). For the \textit{Long Modifier} experiment the $n$-gram shows no few-shot learning (0/8 buckets) and the LSTM shows moderate few-shot generalization (4/8 buckets), but the RNNG and ActionLSTM are still robust (8/8 buckets for both). Across all three experiments, we find that the structurally supervised neural models perform  better than the LSTM ($p<0.001$). 

For these experiments, models were more accurate with transitive verbs, especially for ones that occur infrequently during training. This transitive bias was also present in the \textit{base context} tests for the ActionLSTM and RNNG, indicating that transitvity may be the default assumption, and models expect verbs to be able to occur in passive frames unless they have a large amount of indirect negative evidence to the contrary.

\subsection{Syntactic Invariance} \label{sec:argstruct_invar}

\begin{figure}[]
\begin{minipage}{0.48\textwidth}
\centering
\includegraphics[width=\textwidth]{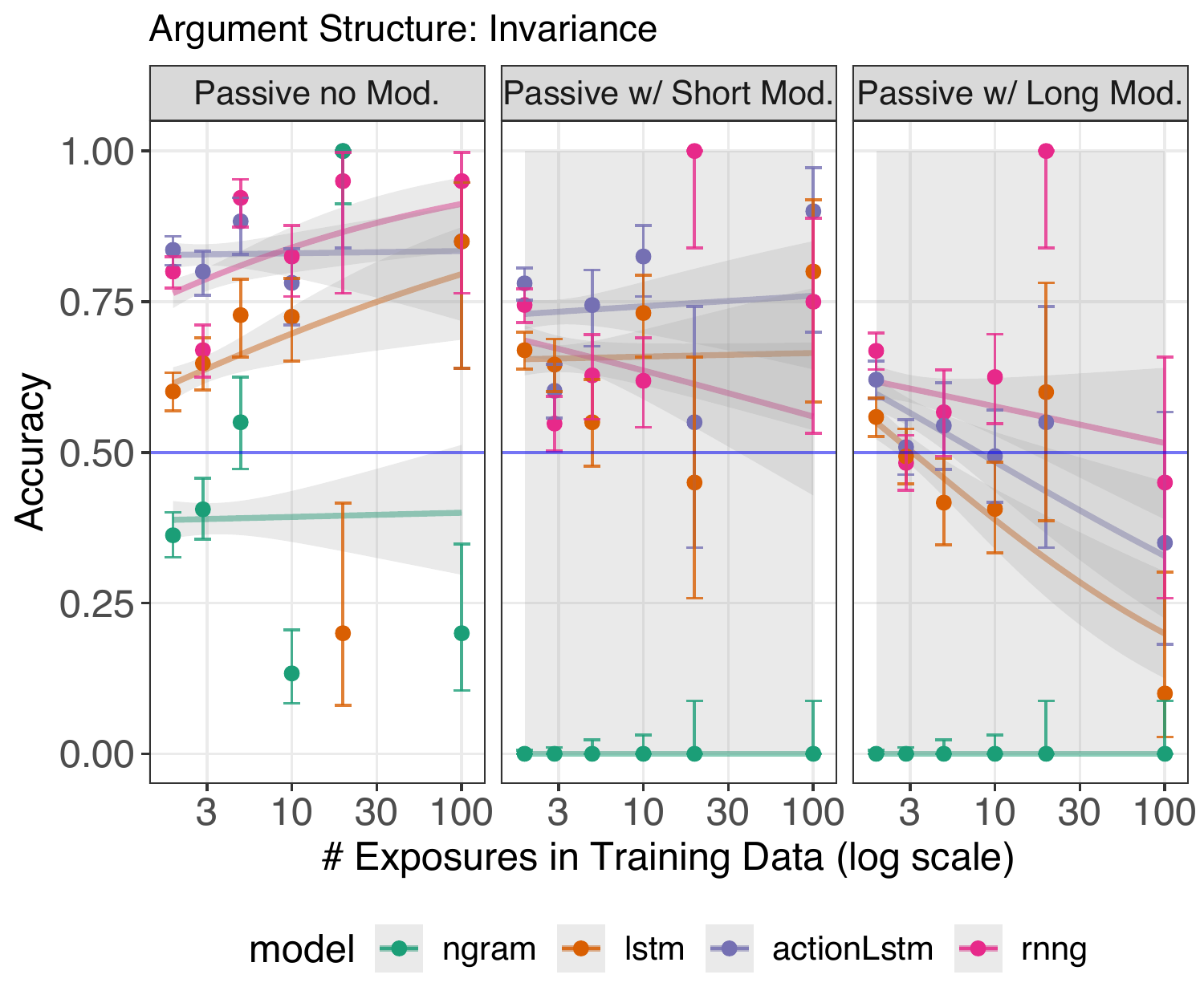}
\end{minipage}
\caption{Invariance to transformation for argument structure. Error bars are 95\% confidence intervals.}
\label{fig:argstruct-invar-results}
\end{figure}

\begin{table*}[!ht]
\small
    \centering
    \begin{tabular}{@{} l M{1.2cm} M{1.2cm} M{1.2cm} M{1.2cm} l M{1.2cm} M{1.2cm} @{}}
        \toprule \multirow{2}{*}[-0.8em]{Test}
        &\multicolumn{4}{c}{Few-Shot Learning} && \multicolumn{2}{l}{Structural Supervision} \\
        \cmidrule{2-5} \cmidrule{7-8}
         & $n$-gram & LSTM & {\footnotesize Action LSTM} & RNNG && {\footnotesize Action LSTM} & RNNG \\
        \midrule
        Number & \cellcolor{yellow}5/8 & \cellcolor{green}8/8 & \cellcolor{green}8/8 & \cellcolor{green} 8/8 && $***$ & $*$ \\
        Number w/ PP Modifier & \cellcolor{red}0/8 & \cellcolor{orange}1/8 & \cellcolor{green}7/8 & \cellcolor{yellow}4/8 && $***$ & $***$  \\
        Number w/ RC Modifier & \cellcolor{red}0/6 & \cellcolor{orange} 2/8 & \cellcolor{green} 7/8 & \cellcolor{yellow} 5/8 && $***$ & $*$ \\
        Verbal Arg. Struct. Infinite & \cellcolor{red}0/6 & \cellcolor{yellow}6/8 & \cellcolor{green}8/8 & \cellcolor{green} 7/8 && $***$ & $***$ \\
        Verbal Arg. Struct. Past & \cellcolor{red} 0/8 & \cellcolor{yellow} 5/8 & \cellcolor{green} 8/8 & \cellcolor{green} 8/8 && $***$ & $***$ \\
        Arg. Struct. Transformation & \cellcolor{green}8/8 & \cellcolor{green}8/8 & \cellcolor{green}8/8 & \cellcolor{green}8/8 && $***$ & $**$  \\
        Arg. Struct. w/ Modifier & \cellcolor{red}0/8 & \cellcolor{green}8/8 & \cellcolor{green}8/8 & \cellcolor{green}8/8 && $*$ & $***$  \\
        Arg. Struct w/ Long Modifier & \cellcolor{red}0/8 & \cellcolor{yellow}4/8 & \cellcolor{green}8/8 & \cellcolor{green}8/8 && $***$ & $***$  \\
        \midrule
        Number Transformation Simple & \cellcolor{red}0/8 & \cellcolor{green}8/8 & \cellcolor{yellow}5/8 & \cellcolor{green}7/8 && ! $***$ & n.s. \\
        Number Transformation w/ Modifier & \cellcolor{red}0/8 & \cellcolor{green}7/8 & \cellcolor{yellow} 4/8 & \cellcolor{green} 7/8 && n.s. & $*$ \\
        Arg. Struct. Transf. & \cellcolor{orange}1/8 & \cellcolor{yellow}5/8 & \cellcolor{green}6/8 & \cellcolor{green}6/8 && $***$ & $***$  \\
        Arg. Struct. Transf. w/ Modifier & \cellcolor{red}0/8 & \cellcolor{yellow}4/8 & \cellcolor{yellow}5/8 & \cellcolor{green}6/8 && $***$ & n.s. \\
        Arg. Struct Transf. w/ Long Modifier & \cellcolor{red}0/8 & \cellcolor{orange}1/8 & \cellcolor{orange}1/8 & \cellcolor{yellow}4/8 && $***$ & $***$ \\
        \bottomrule
    \end{tabular}
    \caption{Left columns: Few shot learning outcomes with the results from our tests of syntactic invariance in the bottom quadrant. Colors correspond to the proportion of exposure buckets for which each model achieved accuracy scores significantly above chance, colored by tertiles. Right columns indicate whether the two structurally supervised models outperform the LSTM for each test, where *s indicate the significance level from our statistical tests and !s indicate significantly worse performance than the LSTM.}
    \label{tab:summary}
\end{table*}

In this section, we run the same \textit{passive} experiments described in Section \ref{sec:argstruct-passive}, however we use verbs that occur only in the \textit{active} voice during training. In order for models to achieve higher than 50\% accuracy, they must learn the co-variation between direct objects in the active voice and passive nominal subjects in the passive voice, even for verbs which they have only seen in the active voice. That is, they must learn a grammatical rule that is invariant to syntactic transformation and verb type.

We sampled all the verbs that occurred with the \texttt{VBD} but not the \texttt{VBN} (past participle) part-of-speech in our training data and generated 20 sentences for each verb. This came to 56 verbs in total, with no transitive verbs in the 50 or 100 exposure buckets. The results from this experiment can be seen in Figure \ref{fig:argstruct-invar-results}. For the \textit{No-Modifier} experiment, the $n$-gram model shows little few-shot learning, above chance in only 1/8 exposure buckets. However, the neural models show moderate few-shot learning, with accuracy scores above chance in 5/8 buckets (LSTM), 6/8 (ActionLSTM) and 6/8 (RNNG). For the \textit{Short Modifier} experiment we find no few-shot learning for the $n$-gram models, but moderate few-shot learning for all neural models tested (LSTM: 4/8 bucekts; ActionLSTM: 5/8 and RNNG: 6/8). Models fare worse in the \textit{Long Modifier} experiment with week few-shot learning for the LSTM and ActionLSTM (1/8 buckets each) and moderate few-shot learning for RNNG model (4/8 buckets each). The $n$-gram is below chance in all buckets. Turning to the effects of structural supervision: We find that the RNNG and the ActionLSTM generally outperform the LSTM ($p<0.001$ for all three, except RNNG/\textit{Short Modifier} which is not significant). Because there were so few verbs in our training data that occurred only in active verbal frames, error estimates are larger for this experiment and the results are somewhat less consistent. Despite this, it is clear that in the \textit{No Modifier} and \textit{Short Modifier} tests, all the neural models show moderate accuracy outcomes, indicating that their learned representations are at least somewhat invariant to syntactic information. 

\vspace{-0.2cm}
\section{Discussion}
\vspace{-0.2cm}

In this paper, we have tested the few-shot learning capabilities of neural language models, as well as whether these models can learn grammatical representations that are invariant to syntactic transformation. First, we addressed neural models' ability to learn \textbf{nominal number}, introducing a novel testing paradigm that leveraged \textit{polar questions} to assess subject/verb number agreement learning in syntactically transformed settings. Second, we turned to neural models' ability to represent verbal \textbf{argument structure}, developing two novel suites of tests that assessed preference for \textit{themes}---either realized as direct objects or passive subjects---in both active contexts and passive contexts. In each experiment we assessed the effect of syntactic supervision on learning outcomes by comparing two supervised models to one purely sequence model.

A summary of our results can be seen in Table~\ref{tab:summary}, with few-shot learning outcomes in colored cells on the left, and the effect of structural supervision on the right. The results from experiments that assess syntactic invariance are on the bottom, below the line break. This table makes it clear that all neural models are capable of making syntactic generalizations about a token from minimal exposure during training. Although model accuracy is reduced for tests that assess syntactic invariance, all neural models show at least a moderate ability to generalize across syntactic transformations. Furthermore, Table~\ref{tab:summary} shows that syntactic invariance is enhanced in structurally supervised models. Interestingly, both ActionLSTM and RNNG have access to syntactic information, but the comparison in Table~\ref{tab:summary} indicates that RNNG can leverage that information more effectively to produce syntactic invariance. Therefore we suggest that RNNG's improved performance does not come from the mere \emph{presence} of syntactic information in the training and test data, but rather from the fact that it uses syntactic information to structure its computation in a non-sequential way.


Because these experiments require careful and robust syntactic analysis of the training data, we evaluated models trained on a relatively small, human-annotated corpus. While the small training data poses some limitations when interpreting the results, it makes them more relevant to low-resource NLP applications and suggests that using structurally supervised models can lead to better generalization in a sparse data environment. While sub-word tokenization schemes such as Byte-Pair Encoding \cite{sennrich2015neural} have helped reduce the number of individual lexical items that need to learned, they do not completely eliminate the long tail of sub-word units. Thus, robust few-shot generalization is still an important problem in these environments. It may be that larger amounts of training data support even better few-shot learning and syntactic invariance outcomes. Scaling these carefully-controlled methods to the larger data setting will be an important next step. However, even with the relatively small models tested here, the results support a growing body of evidence that incremental statistical models of language are able to induce many key features of human linguistic competence.

\vspace{-0.2cm}
\section*{Acknowledgements}
\vspace{-0.2cm}
The authors thank the anonymous reviewers for their feedback. This work was supported by the MIT-IBM Watson AI Lab.

\bibliographystyle{acl_natbib}
\bibliography{anthology,emnlp2020}

\begin{thebibliography}{29}
\expandafter\ifx\csname natexlab\endcsname\relax\def\natexlab#1{#1}\fi

\bibitem[{An et~al.(2019)An, Qian, Wilcox, and Levy}]{an2019representation}
Aixiu An, Peng Qian, Ethan Wilcox, and Roger Levy. 2019.
\newblock Representation of constituents in neural language models:
  Coordination phrase as a case study.
\newblock \emph{arXiv preprint arXiv:1909.04625}.

\bibitem[{Baayen et~al.(1995)Baayen, Piepenbrock, and
  Gulikers}]{baayen1995celex}
R~Harald Baayen, Richard Piepenbrock, and Leon Gulikers. 1995.
\newblock The celex lexical database (release 2).
\newblock \emph{Distributed by the Linguistic Data Consortium, University of
  Pennsylvania}.

\bibitem[{Charniak et~al.(2016)}]{charniak2016parsing}
Eugene Charniak et~al. 2016.
\newblock Parsing as language modeling.
\newblock In \emph{Proceedings of the 2016 Conference on Empirical Methods in
  Natural Language Processing}.

\bibitem[{Chomsky(1957)}]{chomsky1957syntactic}
Noam Chomsky. 1957.
\newblock \emph{Syntactic structures}.
\newblock Walter de Gruyter.

\bibitem[{Dyer et~al.(2016)Dyer, Kuncoro, Ballesteros, and
  Smith}]{dyer2016rnng}
Chris Dyer, Adhiguna Kuncoro, Miguel Ballesteros, and Noah~A. Smith. 2016.
\newblock Recurrent neural network grammars.
\newblock In \emph{Proceedings of the 2016 Conference of the North {A}merican
  Chapter of the Association for Computational Linguistics: Human Language
  Technologies}.

\bibitem[{Elman(1990)}]{elman1990finding}
Jeffrey~L Elman. 1990.
\newblock Finding structure in time.
\newblock \emph{Cognitive science}, 14(2):179--211.

\bibitem[{Futrell et~al.(2018)Futrell, Wilcox, Morita, and
  Levy}]{futrell2018rnns}
Richard Futrell, Ethan Wilcox, Takashi Morita, and Roger Levy. 2018.
\newblock {RNNs} as psycholinguistic subjects: Syntactic state and grammatical
  dependency.
\newblock \emph{arXiv preprint arXiv:1809.01329}.

\bibitem[{Giulianelli et~al.(2018)Giulianelli, Harding, Mohnert, Hupkes, and
  Zuidema}]{giulianelli2018under}
Mario Giulianelli, Jack Harding, Florian Mohnert, Dieuwke Hupkes, and Willem
  Zuidema. 2018.
\newblock Under the hood: Using diagnostic classifiers to investigate and
  improve how language models track agreement information.
\newblock \emph{arXiv preprint arXiv:1808.08079}.

\bibitem[{Gulordava et~al.(2018)Gulordava, Bojanowski, Grave, Linzen, and
  Baroni}]{gulordava2018colorless}
Kristina Gulordava, Piotr Bojanowski, Edouard Grave, Tal Linzen, and Marco
  Baroni. 2018.
\newblock Colorless green recurrent networks dream hierarchically.
\newblock In \emph{Proceedings of the 2018 Conference of the North {A}merican
  Chapter of the Association for Computational Linguistics: Human Language
  Technologies}.

\bibitem[{Hochreiter and Schmidhuber(1997)}]{hochreiter1997long}
Sepp Hochreiter and J{\"u}rgen Schmidhuber. 1997.
\newblock Long short-term memory.
\newblock \emph{Neural computation}, 9(8):1735--1780.

\bibitem[{Hu et~al.(2020)Hu, Gauthier, Qian, Wilcox, and
  Levy}]{hu2020systematic}
Jennifer Hu, Jon Gauthier, Peng Qian, Ethan Wilcox, and Roger~P Levy. 2020.
\newblock A systematic assessment of syntactic generalization in neural
  language models.
\newblock \emph{arXiv preprint arXiv:2005.03692}.

\bibitem[{Jumelet et~al.(2019)Jumelet, Zuidema, and
  Hupkes}]{jumelet2019analysing}
Jaap Jumelet, Willem Zuidema, and Dieuwke Hupkes. 2019.
\newblock Analysing neural language models: Contextual decomposition reveals
  default reasoning in number and gender assignment.
\newblock \emph{arXiv preprint arXiv:1909.08975}.

\bibitem[{Kann et~al.(2018)Kann, Warstadt, Williams, and Bowman}]{kann2018verb}
Katharina Kann, Alex Warstadt, Adina Williams, and Samuel~R Bowman. 2018.
\newblock Verb argument structure alternations in word and sentence embeddings.
\newblock \emph{arXiv preprint arXiv:1811.10773}.

\bibitem[{Kuncoro et~al.(2017)Kuncoro, Ballesteros, Kong, Dyer, Neubig, and
  Smith}]{kuncoro2016recurrent}
Adhiguna Kuncoro, Miguel Ballesteros, Lingpeng Kong, Chris Dyer, Graham Neubig,
  and Noah~A. Smith. 2017.
\newblock What do recurrent neural network grammars learn about syntax?
\newblock In \emph{Proceedings of the 15th Conference of the {E}uropean Chapter
  of the Association for Computational Linguistics}.

\bibitem[{Kuncoro et~al.(2018)Kuncoro, Dyer, Hale, Yogatama, Clark, and
  Blunsom}]{kuncoro2018lstms}
Adhiguna Kuncoro, Chris Dyer, John Hale, Dani Yogatama, Stephen Clark, and Phil
  Blunsom. 2018.
\newblock Lstms can learn syntax-sensitive dependencies well, but modeling
  structure makes them better.
\newblock In \emph{Proceedings of the 56th Annual Meeting of the Association
  for Computational Linguistics (Volume 1: Long Papers)}, pages 1426--1436.

\bibitem[{Lakretz et~al.(2019)Lakretz, Kruszewski, Desbordes, Hupkes, Dehaene,
  and Baroni}]{lakretz2019emergence}
Yair Lakretz, German Kruszewski, Theo Desbordes, Dieuwke Hupkes, Stanislas
  Dehaene, and Marco Baroni. 2019.
\newblock The emergence of number and syntax units in lstm language models.
\newblock \emph{arXiv preprint arXiv:1903.07435}.

\bibitem[{Linzen et~al.(2016)Linzen, Dupoux, and
  Goldberg}]{linzen2016assessing}
Tal Linzen, Emmanuel Dupoux, and Yoav Goldberg. 2016.
\newblock Assessing the ability of {LSTM}s to learn syntax-sensitive
  dependencies.
\newblock \emph{Transactions of the Association for Computational Linguistics},
  4:521--535.

\bibitem[{Marcus et~al.(1993)Marcus, Marcinkiewicz, and
  Santorini}]{marcus1993building}
Mitchell~P Marcus, Mary~Ann Marcinkiewicz, and Beatrice Santorini. 1993.
\newblock Building a large annotated corpus of english: The penn treebank.
\newblock \emph{Computational linguistics}, 19(2):313--330.

\bibitem[{Marvin and Linzen(2018)}]{marvin2018targeted}
Rebecca Marvin and Tal Linzen. 2018.
\newblock Targeted syntactic evaluation of language models.
\newblock In \emph{Proceedings of the 2018 Conference on Empirical Methods in
  Natural Language Processing}.

\bibitem[{Schuster and Manning(2016)}]{schuster2016enhanced}
Sebastian Schuster and Christopher~D Manning. 2016.
\newblock Enhanced english universal dependencies: An improved representation
  for natural language understanding tasks.
\newblock In \emph{Proceedings of the 10th International Conference on Language
  Resources and Evaluation}.

\bibitem[{Sennrich et~al.(2015)Sennrich, Haddow, and
  Birch}]{sennrich2015neural}
Rico Sennrich, Barry Haddow, and Alexandra Birch. 2015.
\newblock Neural machine translation of rare words with subword units.
\newblock \emph{arXiv preprint arXiv:1508.07909}.

\bibitem[{Stern et~al.(2017)Stern, Fried, and Klein}]{stern2017effective}
Mitchell Stern, Daniel Fried, and Dan Klein. 2017.
\newblock Effective inference for generative neural parsing.
\newblock \emph{arXiv preprint arXiv:1707.08976}.

\bibitem[{Stolcke(2002)}]{stolcke2002srilm}
Andreas Stolcke. 2002.
\newblock {SRILM} - an extensible language modeling toolkit.
\newblock In \emph{Proceedings of the 7th International Conference on Spoken
  Language Processing}.

\bibitem[{Tenenbaum and Xu(2000)}]{tenenbaum2000word}
Joshua~B Tenenbaum and Fei Xu. 2000.
\newblock Word learning as bayesian inference.
\newblock In \emph{Proceedings of the Annual Meeting of the Cognitive Science
  Society}.

\bibitem[{Van~Schijndel and Linzen(2018)}]{van2018neural}
Marten Van~Schijndel and Tal Linzen. 2018.
\newblock A neural model of adaptation in reading.
\newblock \emph{arXiv preprint arXiv:1808.09930}.

\bibitem[{Wang et~al.(2017)Wang, Roller, and
  Erk}]{wang-etal:2017-distributional}
Su~Wang, Stephen Roller, and Katrin Erk. 2017.
\newblock Distributional modeling on a diet: One-shot word learning from text
  only.
\newblock In \emph{Proceedings of the 8th International Joint Conference on
  Natural Language Processing}.

\bibitem[{Wilcox et~al.(2018)Wilcox, Levy, Morita, and Futrell}]{wilcox2018rnn}
Ethan Wilcox, Roger Levy, Takashi Morita, and Richard Futrell. 2018.
\newblock What do {RNN} language models learn about filler-gap dependencies?
\newblock In \emph{Proceedings of the 2018 {EMNLP} Workshop {B}lackbox{NLP}:
  Analyzing and Interpreting Neural Networks for {NLP}}.

\bibitem[{Wilcox et~al.(2019)Wilcox, Qian, Futrell, Ballesteros, and
  Levy}]{wilcox2019structural}
Ethan Wilcox, Peng Qian, Richard Futrell, Miguel Ballesteros, and Roger Levy.
  2019.
\newblock Structural supervision improves learning of non-local grammatical
  dependencies.
\newblock In \emph{Proceedings of the 2019 Conference of the North {A}merican
  Chapter of the Association for Computational Linguistics: Human Language
  Technologies}.

\bibitem[{Zipf(1949)}]{zipf1949human}
George~Kingsley Zipf. 1949.
\newblock \emph{Human behavior and the principle of least effort.}
\newblock addison-wesley press.

\end{thebibliography}

\appendix

\section{Effect of Exposure on Model Accuracy}

In this section we report the result for statistical tests assessing the effect of a token's frequency in training on model accuracy for that token. We derive significance from a general linear model with \# of exposures as a sole predictor, with random by-item intercepts (\texttt{glm(accuracy $\sim$ \#\_occurrences + (1|item\_number)))})

\paragraph{Nominal Number} For the base context, in the no modifier condition we find a positive effect of increased exposure for all models ($p<0.001$). For the \textit{PP modifier} test we find an effect of exposure for the ActionLSTM and the RNNG ($p<0.001$), and a negative, but insignificant effect for the $n$-gram and the LSTM. For the \textit{RC Modifier} experiment we find an effect of increased exposure for all three neural models ($p<0.001$ for the RNNG and ActionLSTM; $p<0.05$ for the LSTM), but no effect for the $n$-gram. For the inverted contexts: in the \textit{no modifier} tests we find no effect of increased exposure, except for the LSTM, where the effect is negative ($p<0.01$). For the \textit{modifier} tests, we find a significant effect for the ActionLSTM and the RNNG ($p<0.001$).

\paragraph{Argument Structure} For the base context (active voice): In the infinitival tests, we find a significant effect of exposure on accuracy for the ActionLSTM and the RNNG ($p<0.001$) and a negative effect for the $n$-gram model ($p<0.001$). In the past-tense, we find no significant effect for the RNNG or the ActionLSTM, and a negative effect for the $n$-gram and LSTM models ($p<0.001$). In the transformed contexts (passive voice), for the \textit{no-modifier} tests we find a significant effect of exposure for all models ($p<0.001$ for all except ActionLSTM where $p<0.05$). For the \textit{short-modifier} tests we find an effect for the ActionLSTM ($p<0.05$) and the RNNG ($p<0.001$). And in the \textit{long-modifier} test we find a marginally significant effect for the three neural models ($p\sim 0.05$ for all).

\section{Learning Outcomes by Grammatical Condition}

\begin{figure}[]
\begin{minipage}{0.48\textwidth}
\centering
\includegraphics[width=\textwidth]{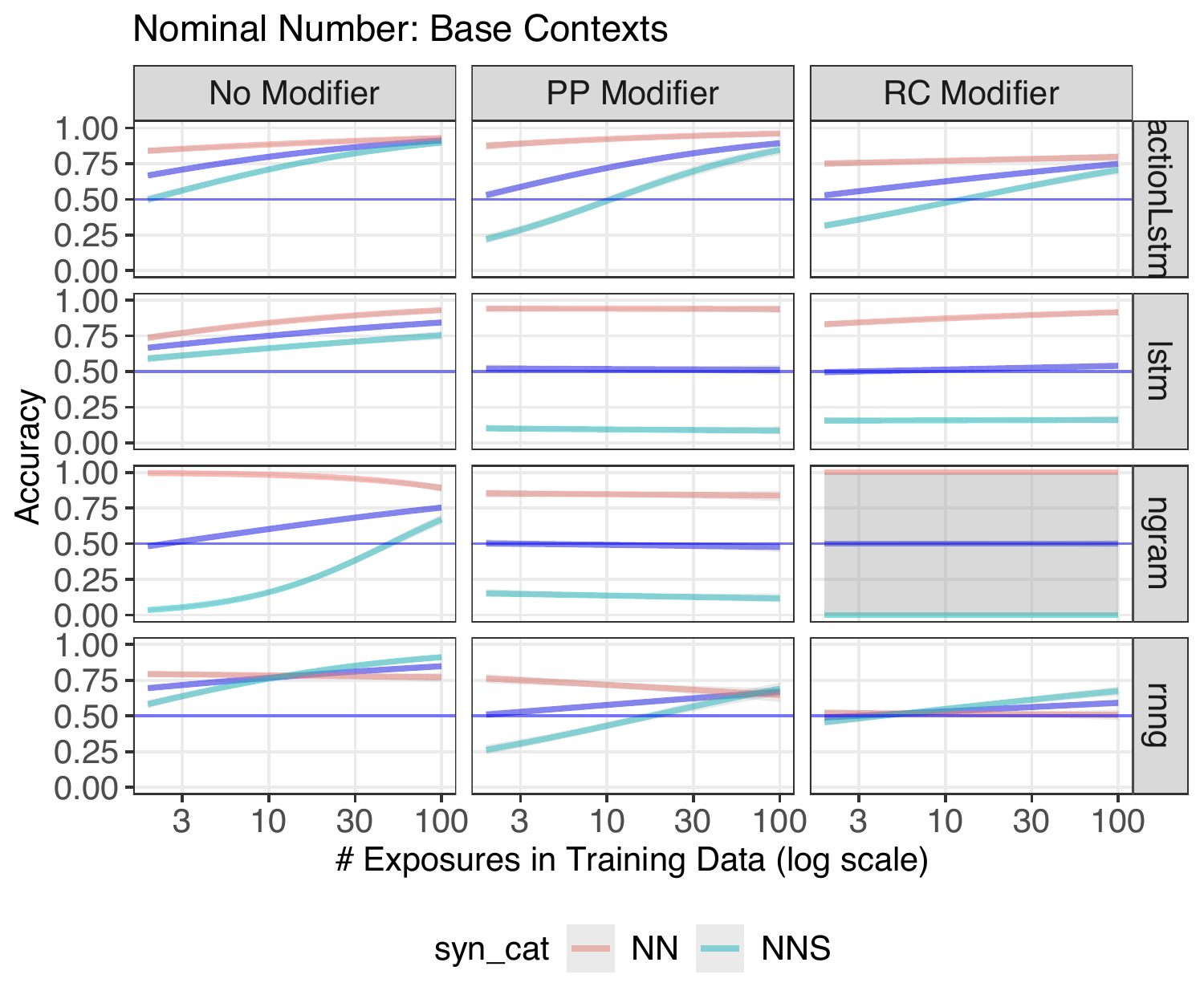}
\caption{Number: Base Contexts}
\label{fig:num-base-appendix}
\end{minipage}
\end{figure}

\begin{figure}[]
\begin{minipage}{0.48\textwidth}
\centering
\includegraphics[width=\textwidth]{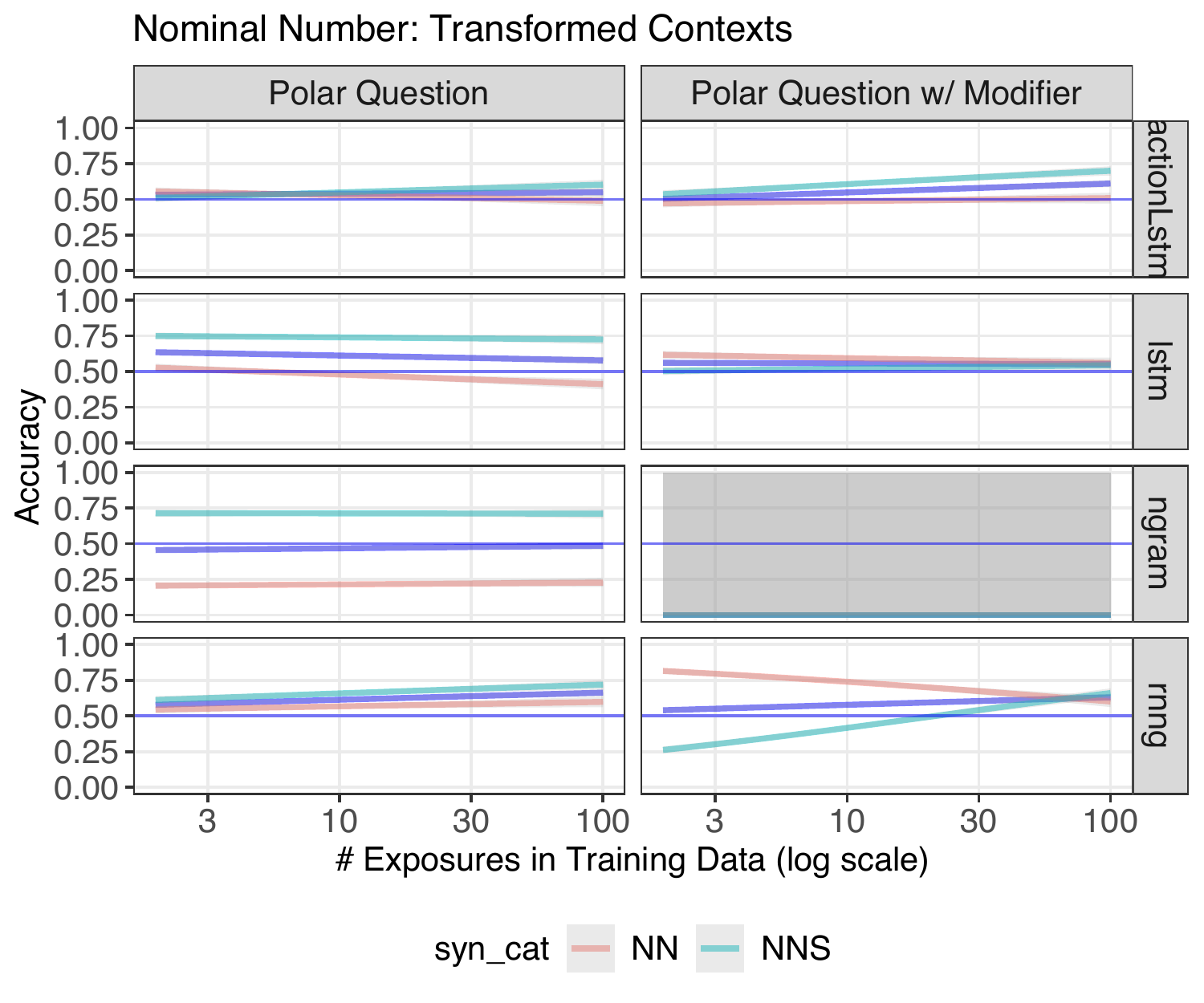}
\caption{Number: Transformed Contexts}
\label{fig:num-transf-appendix}
\end{minipage}
\end{figure}

In this section, for each test reported in the paper, we break down model performance by grammatical category, either singular vs plural nouns (for nominal number tests) or transitive vs. intransitive verbs (for our argument structure tests). Charts follow the same presentational paradigm: $y$-axis shows accuracy and $x$-axis the number of times each word appears during training, on a log-10 scale. Smooth lines are results of logistic regression model fits on the raw data, with shaded regions indicating standard error. Dark blue lines show model performance averaged between the two conditions (these are the same fits that appear in charts in the main body of the paper). 

The data presented here are consistent with the hypothesis from \cite{jumelet2019analysing}. When models receive scant evidence of a token's syntactic properties in training, they assume that it belongs to a ``base" category, which is \textit{singular} for nouns and \textit{transitive} for verbs. Thus, models are more accurate for singular nouns and transitive verbs seen rarely in training. As the model receives more evidence that a token is not in the base category, its predictions flip. Hence, gains in overall-accuracy tend to come from models learning the proper agreement for non-base tokens (plural nouns and intransitive verbs). Generally, these effects are stronger for nominal number learning, and stronger for structurally supervised models than for the LSTM, which is consistent with the findings presented in the main body of the text.

\subsection{Number: Base Contexts}

The nominal number breakdown for base contexts can be seen in Figure \ref{fig:num-base-appendix}, with accuracy scores for singular nouns (\texttt{NN}) in red and plural nouns \texttt{NNS}) in teal. Over all, models tended to show higher accuracy scores for singular nouns, which indicates the presence of a singular bias. Interestingly, the ActionLSTM and the RNNG are capable of overcoming the singular bias when presented with sufficient data, however the LSTM remains equally biased for tokens seen 2 and 100 times in training.

\subsection{Number: Transformed Contexts}

\begin{figure}[]
\begin{minipage}{0.48\textwidth}
\centering
\includegraphics[width=\textwidth]{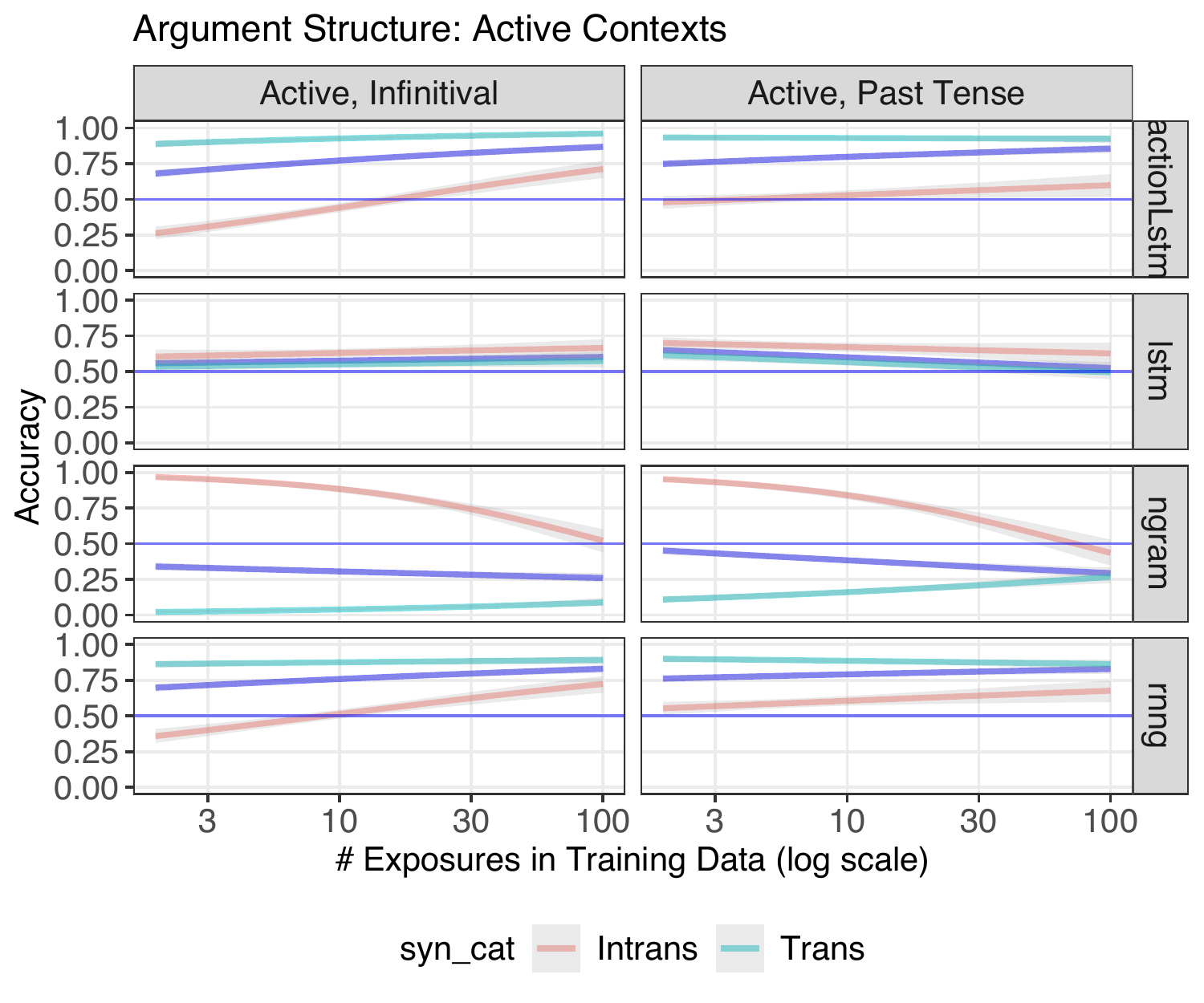}
\caption{Number: Arg Struct: Base Contexts}
\label{fig:arg-base-appendix}
\end{minipage}
\end{figure}

\begin{figure}[]
\begin{minipage}{0.48\textwidth}
\centering
\includegraphics[width=\textwidth]{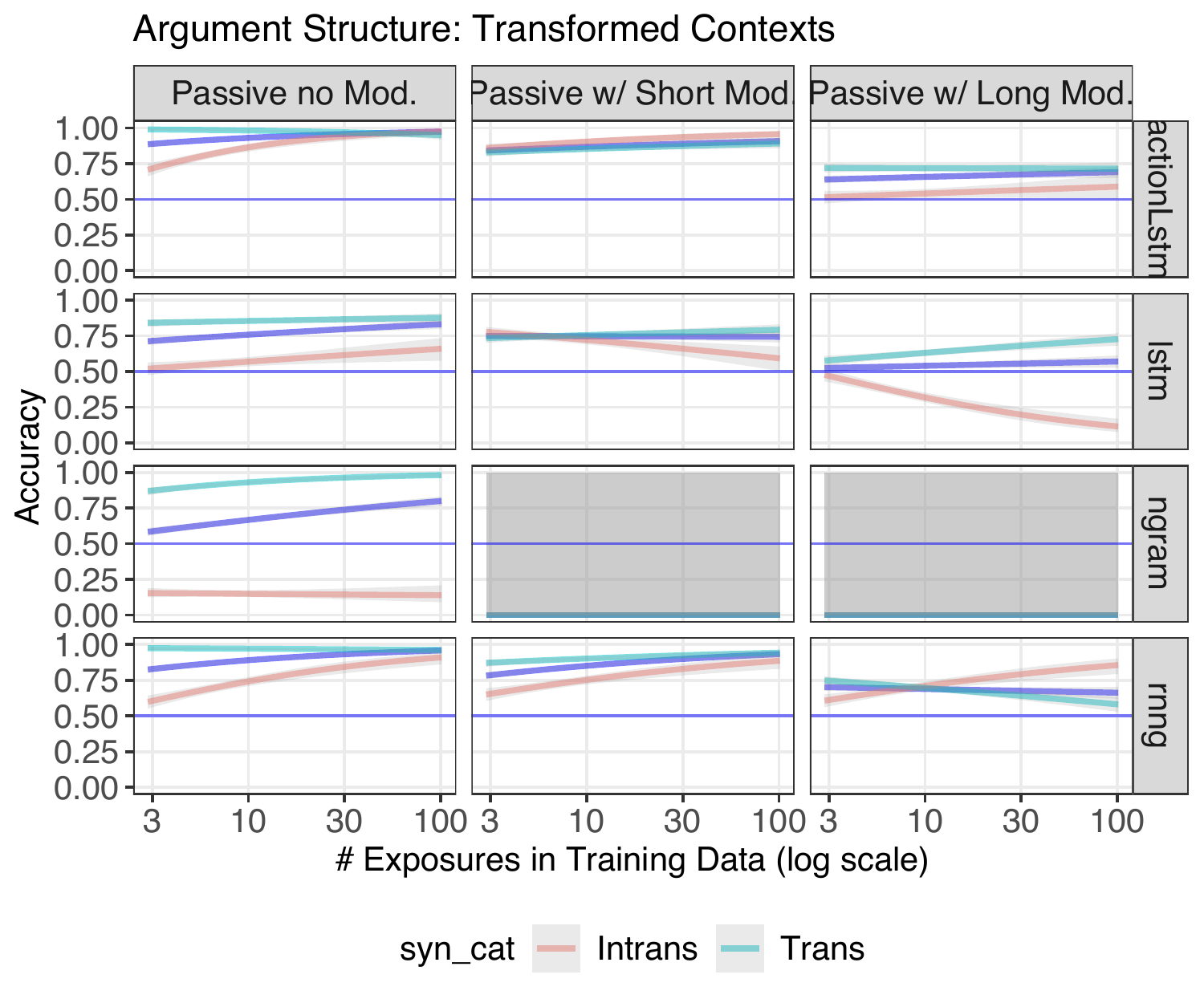}
\includegraphics[width=\textwidth]{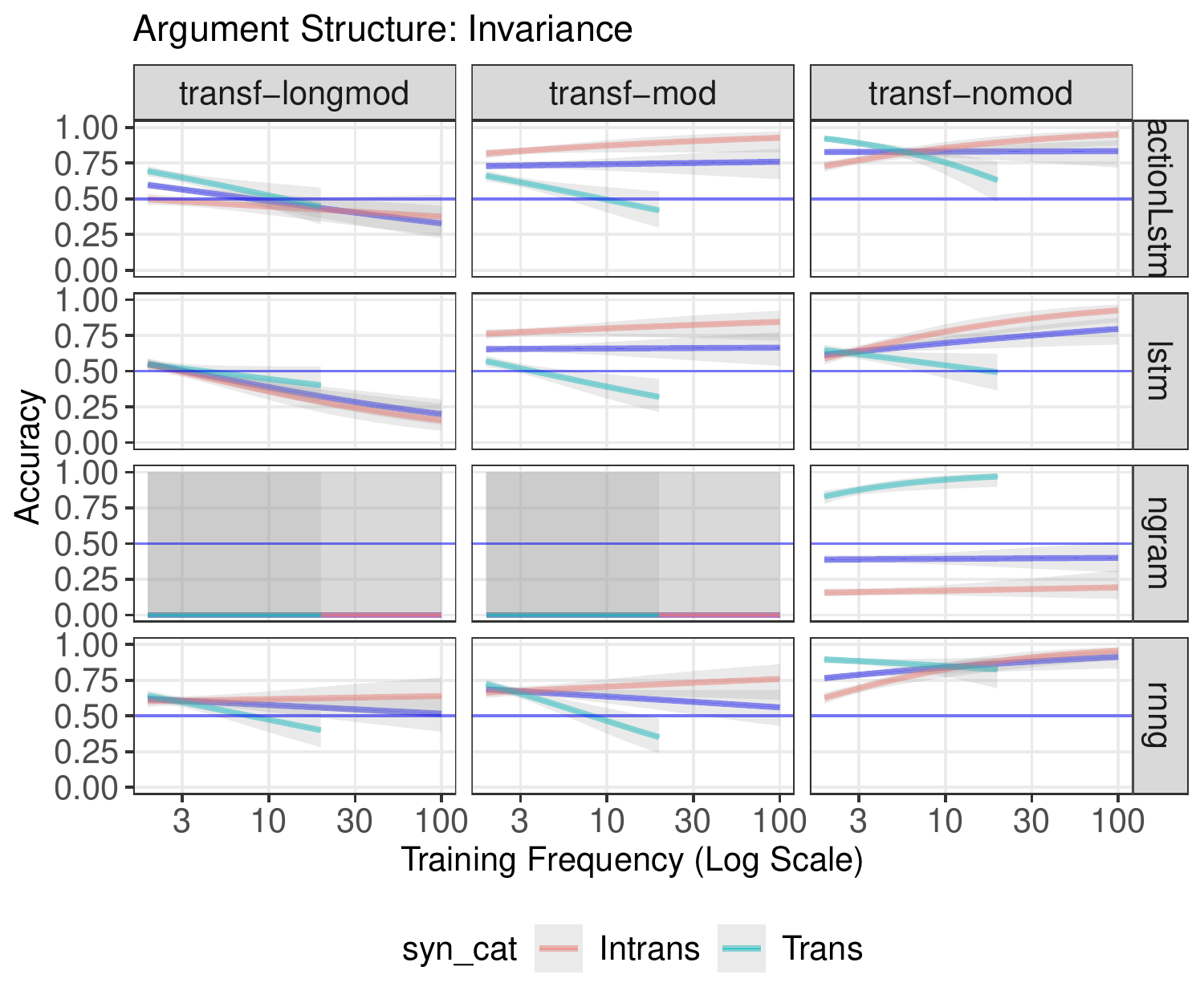}
\caption{Argument Structure: Transformed Contexts (top) and Invariance (bottom) }
\label{fig:arg-transf-appendix}
\end{minipage}
\end{figure}

The nominal number breakdown for transformed can be seen in Figure \ref{fig:num-transf-appendix}. The empirical picture is more complicated here, however if anything models show higher performance for plural nouns. This behavior suggests that \textit{is} sets up weaker expectations for singular nouns than \textit{are} does for plural nouns. Such a pattern is consistent with the hypothesis that models learn the singular as a base form, in which case it would set up weaker expectations for singular nouns. These results compliment those from \citet{an2019representation} (section 6), who also test in inverted settings and find that models tend not to be surprised at coordinated NPs following a singular verb, as in the ungrammatical sentence \textit{*What is the pig and the cat eating?}

\subsection{Argument Structure: Base Contexts}

The breakdown for argument structure learning base contexts can be seen in Figure \ref{fig:arg-base-appendix}, with accuracy scores for \textit{intransitive} verbs in red and \textit{transitive} verbs in teal. Here, we see a strong transitive bias for the two structurally supervised models, with no obvious bias for the LSTM and an intransitive bias for the $n$-gram. 

\subsection{Argument Structure: Transformed Contexts and Invariance}

The breakdown for argument structure learning in the transformed contexts can be seen in Figure \ref{fig:arg-transf-appendix} with transformation tests on the top and invariance tests on the bottom. In this case, where performance is different between the two conditions models display higher accuracy scores for transitive verbs.

\end{document}